\documentclass[letterpaper, 10 pt, conference]{ieeeconf}  
\IEEEoverridecommandlockouts                              

\usepackage{graphics} 
\usepackage{amsmath} 
\usepackage{amssymb}  
\usepackage{siunitx}
\usepackage{mathtools}
\usepackage{booktabs}
\usepackage[ruled]{algorithm2e}
\usepackage{subcaption} 
\usepackage{adjustbox}
\usepackage{pbox}
\usepackage{dblfloatfix}
\usepackage{graphicx}
\usepackage{xcolor}
\usepackage{hyperref}
\usepackage{caption}
\usepackage{subcaption}
\usepackage{cite}
\usepackage{mathrsfs}
\usepackage{url}
\usepackage{multirow}
\usepackage[bottom=57pt,top=54pt, left=54pt, right=54pt]{geometry}   

\newcommand{\etal}{et~al.}
\DeclareMathAlphabet{\pazocal}{OMS}{zplm}{m}{n}

\begin{document}

\title{
Fast and Compute-efficient Sampling-based Local Exploration Planning via Distribution Learning
}


\author{Lukas Schmid$^{\ast}$, Chao Ni$^{\ast}$, Yuliang Zhong,  
Roland Siegwart, and Olov Andersson

\thanks{$^\ast$ Authors contributed equally.}
\thanks{This work was supported by funding from the Microsoft Swiss Joint Research Center and a Wallenberg Foundation and WASP Postdoctoral Scholarship.}
\thanks{The authors are with Autonomous Systems Lab, Department of Mechanical and Process Engineering, ETH Z\"urich, Z\"urich, Switzerland.}%
 \thanks{
{\tt\small schmluk@ethz.ch}}%
}

\maketitle

\begin{abstract}
Exploration is a fundamental problem in robotics. While sampling-based planners have shown high performance and robustness, they are oftentimes compute intensive and can exhibit high variance. 
To this end, we propose to learn both components of sampling-based exploration. We present a method to directly learn an underlying informed distribution of views based on the spatial context in the robot's map, and further explore a variety of methods to also learn the information gain of each sample.
We show in thorough experimental evaluation that our proposed system improves exploration performance by up to 28\% over classical methods, and find that learning the gains in addition to the sampling distribution can provide favorable performance vs. compute trade-offs for compute-constrained systems.
We demonstrate in simulation and on a low-cost mobile robot that our system generalizes well to varying environments.
\end{abstract}


\section{Introduction}
\label{sec:introduction}

Exploration of unknown environments is a fundamental problem in robotics. It is essential for a variety of applications, ranging from consumer robotics to autonomous surveying and inspection, as well as search and rescue. 
In this paper, we consider the problem of exploring unknown environments represented by standard occupancy maps, encoding free, occupied and unknown space. This common map representation can be obtained from any volumetric mapping system~\cite{hornung2013octomap, oleynikova2017voxblox, schmid2021panoptic} and enables a ground robot to plan in continuous 2D space.

In recent years, a clear trend has emerged to split the exploration problem into the distinct sub-problems of local and global exploration~\cite{history_nbvp, meng_2stage_expl, dang2019gbplanner, Selin_nbv_fron, Schmid21Glocal}.
This can be explained by the conflicting objectives of complete coverage and fast exploration progress, which can be tackled independently in local and global planning.
In local exploration, the goal is to generate a safe path that uncovers as much of yet unmapped space in as little time as possible. Complementary, global exploration aims at identifying a traversal order of frontiers, the boundaries between known and unknown space, to escape local minima and achieve complete coverage.
In this work, we focus solely on the problem of local exploration.

In similar fashion, two main families of approaches, being frontier and sampling-based methods, have emerged.
In frontier-based exploration~\cite{yamauchi1997frontier, cieslewski2017rapid} the frontiers are directly used to compute exploration goals. In the other, sampling-based methods first, i) sample viewpoints in the mapped space, and ii) assign a utility value to views, to identify the Next-Best-View (NBV)~\cite{history_nbvp, meng_2stage_expl, dang2019gbplanner, Selin_nbv_fron, Schmid21Glocal, nbvp_bircher, Schmid20ActivePlanning, dai2020fast, song2018surface}. 

Although there exist extensions of frontier-based methods for rapid flight~\cite{cieslewski2017rapid}, sampling-based planning has found significantly more success in local exploration~\cite{history_nbvp, meng_2stage_expl, dang2019gbplanner, Selin_nbv_fron, Schmid21Glocal, nbvp_bircher, Schmid20ActivePlanning, dai2020fast}. 
This is primarily due to their ability to capture volumetric information for each view point, the possibility to account for varying information gain objectives~\cite{Schmid20ActivePlanning, song2018surface}, and trade-off gains and costs for each viewpoint~\cite{history_nbvp, meng_2stage_expl, dang2019gbplanner, Selin_nbv_fron, Schmid21Glocal, nbvp_bircher, Schmid20ActivePlanning, dai2020fast, song2018surface}.

\begin{figure}[t]
    \centering
    \includegraphics[width=\linewidth]{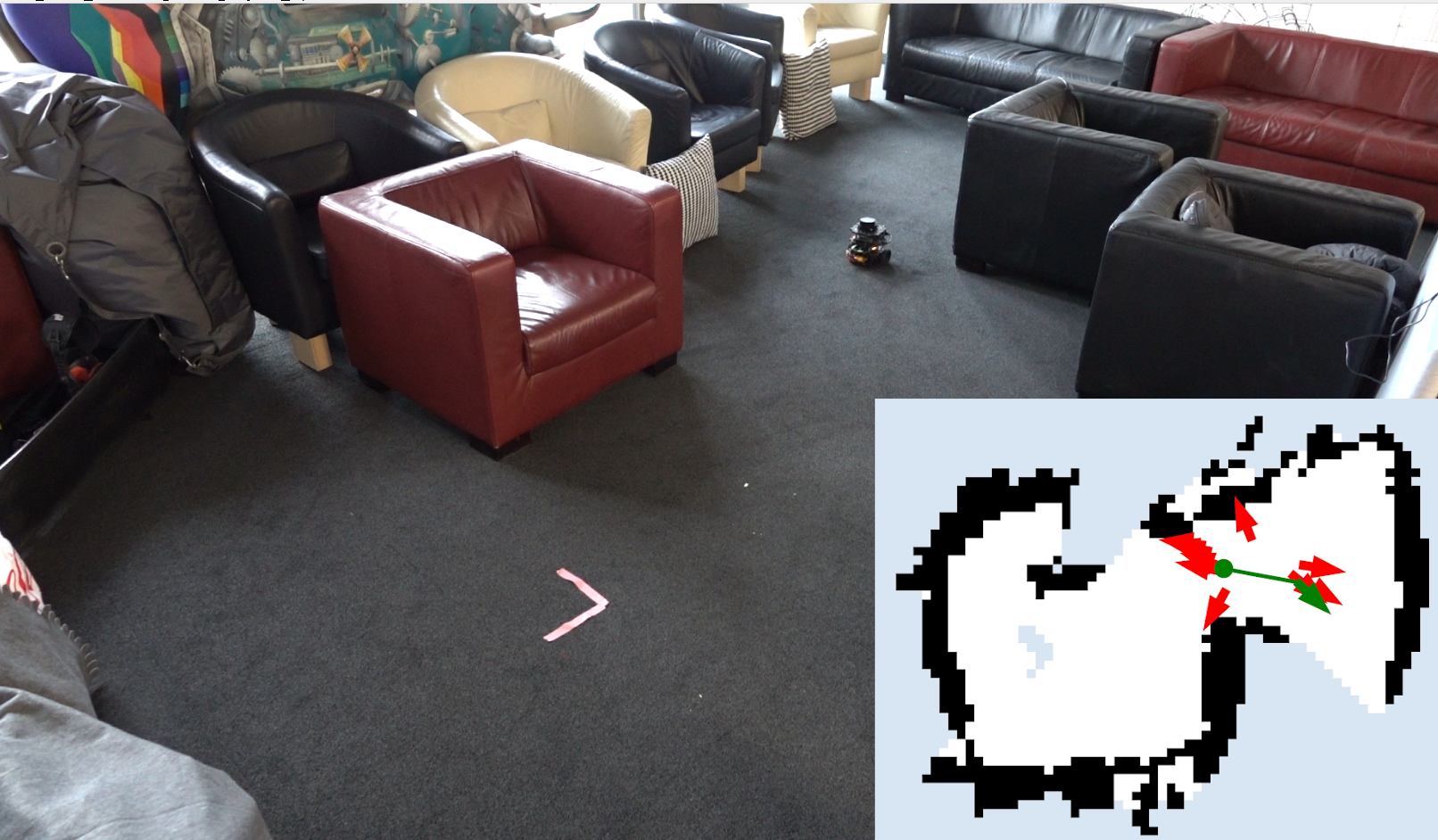}
    \caption{A small compute-constrained mobile robot exploring an office environment, overlaid with its built occupancy map. Our proposed method uses the map to generate an informed distribution viewpoints (red arrows), capturing the inherently multi-modal nature of exploration planning reflecting the regions of gray unknown space. It identifies the NBV (green) to quickly explore the scene.}
    \label{fig:intro_showcase}
    \vspace{-20pt}
\end{figure}

However, traditional sampling-based methods also have notable shortcomings.
First, they are oftentimes highly compute intensive, potentially reducing performance on compute constrained mobile robots, and leaving less capacity for other tasks such as mapping or scene understanding.
Second, the performance of such methods is strongly governed by the number of samples and the randomness inherent to the sampling procedure.
Therefore, many samples may be needed to identify good paths, such as e.g. when most of the space in a room is already explored, or the robot has to traverse a narrow passage.

Simultaneously, learning-based methods have shown strong potential in robotics. 
Typically, active planning problems are tackled via Reinforcement Learning (RL) \cite{babaeizadeh2017ga3c, henderson2018deep, zhu2018, niroui2019, chen2019self} or imitation learning \cite{andersson2017deep, reinhart2020, Bai2017}, learning a mapping of the current state to an action. 
However, RL methods are known to be hard to train \cite{henderson2018deep} and typically require significant simplifications. 

To overcome these limitations, we propose a learning-based approach that follows the sampling-based exploration paradigm. 
To this end, we propose to learn an underlying informed distribution of valuable view points, accounting for information gain, kinematic feasibility, collision avoidance, and dynamics-based motion cost, directly from the robot map.
We further explore different variations of learning to efficiently estimate the information gain of a sample, reducing the computational requirements.
We show in thorough experimental evaluation in simulated and real world experiments that this approach is interpretable and robust, simultaneously improving exploration performance while reducing computational cost, generalizing to different environments.

We make the following contributions:
\begin{itemize}
    \item We propose a new approach to local exploration by learning the key components of sampling-based exploration planning.
    \item We propose a method based on conditional variational autoencoders (CVAE) to directly learn informed multi-modal sampling distributions for exploration planning from standard occupancy maps. We show that this approach is interpretable and achieves superior performance compared to conventional exploration planners and an imitation learning baseline.
    \item We further explore a variety of approaches to learn efficient estimates of the expensive gain computation, and find that CVAE-sampling combined with learned gains offers favorable compute vs. performance trade-offs for compute constrained systems.
    \item We thoroughly validate the generalization capability of our approach with both simulated and real-world experiments. Our approach transfers to a small mobile robot with low-cost sensors and maintains good performance without tuning or retraining. We make our code available as open-source\footnote{\url{https://github.com/ethz-asl/cvae_exploration_planning}}.
\end{itemize}
\section{Related Work} 
\label{sec:rel_work}

\subsection{Sampling-based Local Exploration Planning}
\label{sec:rel_expl}

Sampling-based planning has found distinct success in exploration planning. To find the NBV, typically viewpoints are sampled, information gains and final utilities of each sample are computed, and the best view is selected.
Bircher~\etal~\cite{nbvp_bircher} store samples in a Rapidly-exploring Random Tree (RRT) structure to create a sequence of traversable poses. The gain of each viewpoint is the number of observable voxels, discounted by an exponential distance term.
This method has been widely used and extended in subsequent works \cite{history_nbvp, meng_2stage_expl, dang2019gbplanner, Selin_nbv_fron}. 
However, these mostly innovate on the global planner and use similar local planners to \cite{nbvp_bircher}.
An alternative to distance-based discounts is to normalize the gain by the cost \cite{Schmid20ActivePlanning, Schmid21Glocal}, which is naturally closer to the final exploration objective. 
While \cite{Schmid20ActivePlanning} uses this utility to create informative long-term plans for different applications, \cite{Schmid21Glocal} shows that, for exploration, the majority of the relevant data lies within a very short planning horizon and superior performance can be achieved via local planning.
All of the above methods use uniform random sampling.
This is motivated by the advantage of uniform sampling being asymptotically complete, and it typically achieves good coverage in practice. 

A different line of approaches attempt to manually engineer heuristics for informed sampling in exploration \cite{dai2020fast, song2018surface, kompis2021informed}, based on computing the location of frontiers and sampling points around these as a heuristic for where informative poses may lie. However, this can lead to poses that are not easily accessible and come at an increased computational cost.
As engineering informed and computationally cheap heuristics is generally a non-trivial problem, we argue that it can instead be learned automatically, where the full contextual information in the map can be used to inform NBV sampling.

\subsection{Learning for Path Planning and Exploration}
\label{sec:rel_learning}
With recent successes of machine learning across multiple domains, leveraging learning approaches for motion planning problems has also been receiving increasing attention. However, most works so far consider the classical motion planning problem with a known map and a given goal \cite{lmp:ChenDLYLS20,lmp:quershi2021,lmp:wang2020}. Ichter et al. \cite{ichter2018learning} also learns a sampling distribution, but only considers classical motion planning with known maps and uses a high-level representation of the environment (e.g. object positions) instead of standard occupancy maps.
Some works relax the assumption of a known map but they still focus on finding a path to a given goal \cite{MirowskiPVSBBDG17, zhelo2018curiosity}.

The problem of learning to explore unknown environments has received less attention. 
RL approaches are in theory the most general, but as they can be difficult to train \cite{henderson2018deep}, these tend to rely on significant simplifications. Zhu et al. \cite{zhu2018} consider the problem of exploring office environments and use an actor-critic RL approach to act as a filter for a classical planner, selecting one of six sectors to sample from. Similarly, Niroui et al. \cite{niroui2019} first extract frontiers and use an RL approach to select between them. Chen et al. \cite{chen2019self} used RL with a pre-defined quasi-random distribution of actions for a robot with omnidirectional sensor and kinematics.

Another learning approach popular for planning and control tasks is to use supervised learning to imitate a conventional planner, which can be computationally cheaper \cite{andersson2017deep}. In the area of exploration planning, Reinhart et al. \cite{reinhart2020} learn to imitate a planner tailored to tunnel environments, where the action space is limited to trajectories in eight regions with a fixed length of \SI{2}{\meter}. Englot et al. \cite{Bai2017} similarly discretize the action space into 36 angles on a circle around the robot and show a reduction in computational cost over exhaustively evaluating the gains for all such actions. 
While learning holds great potential, current approaches do not allow sampling viewpoints freely in space, instead relying on pre-defined action distributions and often holonomic robots with omnidirectional sensors. Such assumptions do not always well reflect the constraints of real robots. 
We thus propose to combine the advantages of learning and sampling-based planners, learning both an informed sampling distribution from the robot map, as well as the information gain of each sample, to efficiently compute the utility and NBV.

\section{Problem Statement}
\label{sec:problem}

\begin{figure}[tb]
    \centering
    \includegraphics[width=0.8\linewidth]{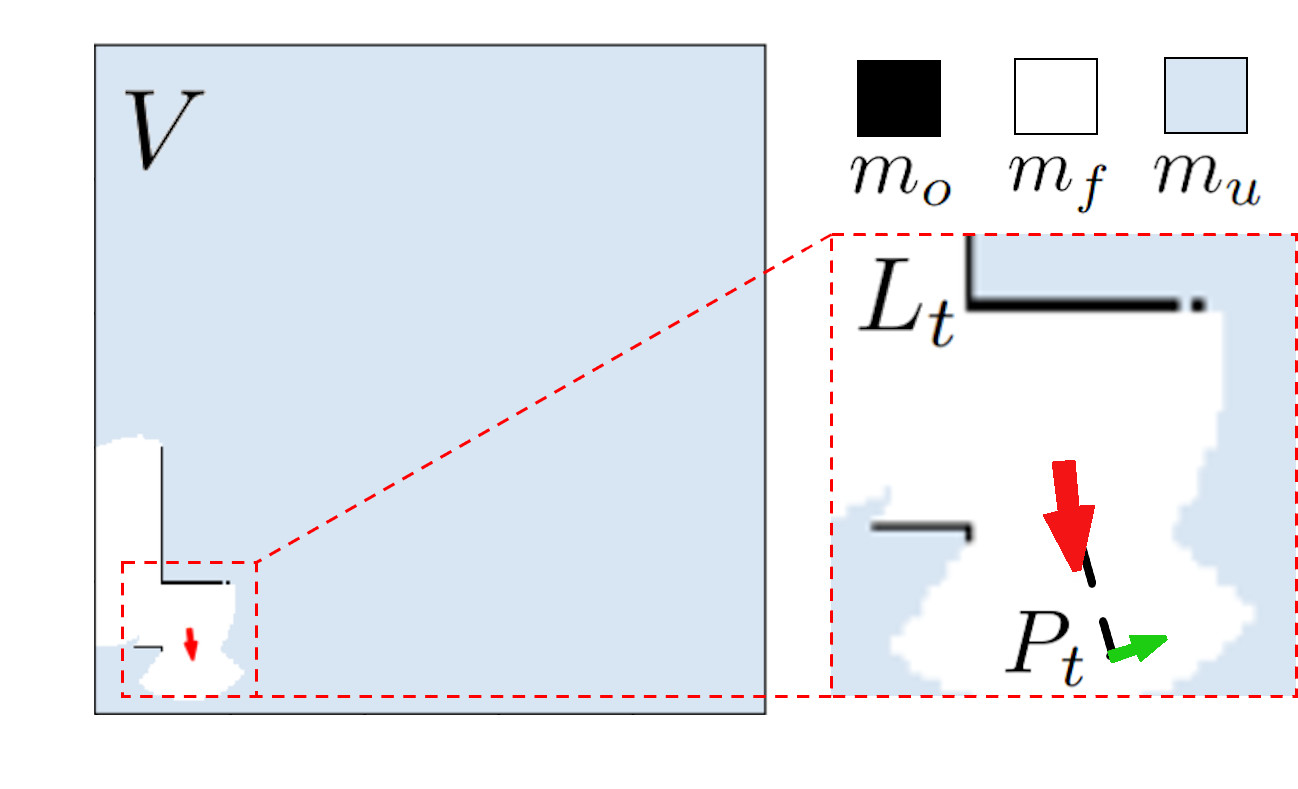}
    \vspace{-10pt}
    \caption{Problem formulation. Given a local map $L_t \subseteq V$, $L_t(v):\mathbb{R}^2 \mapsto \mathbb{M} = \{m_o, m_f, m_u\}$ being occupied, free, and unknown areas. The robot (red) then has to compute a NBV pose $P_t$ (green). }
    \label{fig:problem_statement}
    \vspace{-20pt}
\end{figure}

In this paper, we address the problem of using learning to improve sampling-based local exploration 
using conventional occupancy maps, visualized in Fig.~\ref{fig:problem_statement}.

\textbf{World Representation} 
Given is an area of interest represented as a voxel grid $V\subseteq \mathbb{R}^2$. Each voxel $v \in V$ has a state in the robot map at time $t$ $M_t(v): \mathbb{R}^2 \mapsto \mathbb{M} = \{m_o, m_f, m_u\}$, denoting occupied, free, and unknown, respectively. Initially $M_0(v) = m_u\; \forall v \in V$. Here $V_o \subseteq V$ denotes the observable space for a given robot.

\textbf{NBV Planning}
At each time step $t$, the robot is given a local map of the environment $L_t \subseteq V$. The task is then to identify the NBV pose $P_t = \{x_t,y_t,\theta_t\} \in \mathbb{R}^2 \times SO(2)$ within $L_t$, where the robot moves to.

\textbf{Objective Function}
The final goal of a planning mission can be formulated as a constrained optimization problem:
\begin{align}
    \label{eq:problem_formulation}
    P_0^\ast,\dots, P_{N_P}^\ast = \min_{P_0,\dots,P_{N_P}} \sum_{i=0}^{N_P} T(P_i) + \gamma C(P_i)\\
    \text{subject to } M_{N_P}(v) \neq m_u,\; \forall v \in V_o,
\end{align}
for the first to last selected views $P_0 \dots P_{N_P}$, where $T(P_i)$ indicates the time to \textit{move} the robot from $P_{i-1}$ to $P_i$, and $C(P_i)$ indicates the time it takes to \textit{compute} the next goal pose. Here $\gamma$ allows trading off these objectives to select a planning approach most suited to the hardware and task.


\section{Approach} 
\label{sec:approach}

\begin{figure}[tb]
    \centering
    \includegraphics[width=\linewidth]{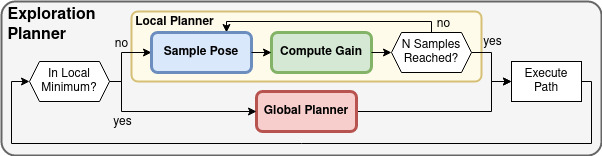}
    \caption{System overview. We focus on sampling-based local planning, which can be split into sampling poses and computing the view-gains. A standard global planner is used if the robot is in a local minimum.}
    \label{fig:overview}
    \vspace{-20pt}
\end{figure}

An overview of our planning system is shown in Fig.~\ref{fig:overview}, adopting the two-stage global-local planning approach \cite{history_nbvp, meng_2stage_expl, dang2019gbplanner, Selin_nbv_fron, Schmid21Glocal}. 
This work focuses on improving local sampling-based exploration, which can be split into the sampling of candidate poses and computation of an information gain for each candidate. Both of these components are detailed in the sections below. 

When the planner reaches a local minimum $\nexists v \in L_t | M_t(v) = m_u \wedge \mathrm{reachable}(v)$, i.e. there is no newly observable voxel left in the local robot map, or the planner has not observed any voxel in 5 consecutive actions, the global planner is called.
We use a frontier-based global planner similar to \cite{Schmid21Glocal} to move the robot to the closest unexplored site and continue local planning.

\subsection{Learning Local Sampling Distributions}

\begin{figure*}[tb]
    \centering
    \includegraphics[width=\linewidth]{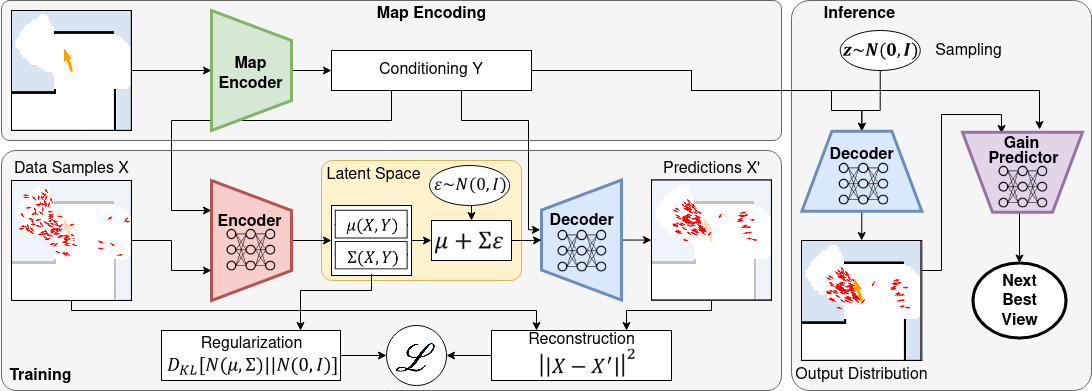}
    \caption{System architecture. The encoder and decoder are trained offline in CVAE fashion to learn an underlying distribution view point for a map, given demonstrations of informative samples. The map encoder and gain predictor can be trained independently in supervised fashion. At run-time, any random seed $z$ can be drawn from the prior distribution and mapped to an output sample. After a gain is assigned and utility computed, the next action is chosen.}
    \label{fig:architecture}
    \vspace{-20pt}
\end{figure*}
The central idea of our approach is based upon the observations, that i) in exploration planning the distribution of informative viewpoints is typically multi-modal and ii) the robot map can provide significant spatial context.
We thus propose to directly learn an informed distribution, accounting for information gain, kinematic feasibility, collision avoidance, and dynamics-based motion cost, for an environment $L_t$. Since the set of possibles poses $P'_t = \{x_t, y_t, \theta_t\} \in L_t$ quickly becomes intractable to optimize over, we take advantage of sampling-based planning to generate a finite number of high utility view points, detailed in Sec.~\ref{sec:data}.
We then use these poses as a supervision signal to learn an underlying informed distribution.
In particular, we leverage Conditional Variational Auto-Encoders (CVAE) to capture a hidden latent distribution \cite{sohn2015cvae} based on training samples. An overview of our architecture is shown in Fig.~\ref{fig:architecture}. 

\textbf{Preliminaries}
We denote $X$ the target variable, parametrized as pose samples $P \in \mathbb{R}^2 \times SO(2)$, $X = \{P_i\}_{i=0,\dots,N_X}$. We denote the conditioning variable as $Y$ and the latent representation as $Z$.
As the true distribution $p(Z|X,Y)$ is unknown, it is approximated by a decoder $Q_\phi(Z|X,Y)$, where $Q$ is parametrized as a neural network with parameters $\phi$. Similarly, the generative distribution $P_\psi(X|Z,Y)$ is modelled as the decoder network with parameters $\psi$.
We model the hidden latent distribution  $p(Z|Y)$ as Gaussian, and use the reparametrization trick to propagate gradients through the sampling step of $Z$. The non-linearity of $Q_\phi$ and $P_\psi$ is leveraged to map $Z$ to the multi-modal distribution of $X$.
As the true distribution $p(X|Y)$ is intractable, we optimize the variational lower bound $\mathscr{L}\leq \log p(X|Y)$: \vspace{-15pt}

\begin{equation}
\resizebox{1.0\columnwidth}{!}{$\mathscr{L} = \mathbb{E}_{Z\sim Q_\phi}[\log P_\psi(X|Z,Y)-D_{KL}(Q_\phi(Z|X,Y)\Vert p(Z|Y))]$} \label{eq:elbo}
\vspace{-5pt}
\end{equation}

where $D_{KL}(\cdot)$ is the Kullback-Leibler divergence. $\mathscr{L}$ can be approximated by Monte Carlo methods and efficiently optimized by standard learning techniques \cite{doersch2016tutorial, kingma2013auto}.

\textbf{Map Encoding}
Since the robot environment is represented by a high dimensional occupancy map $L_t \in \mathbb{M}^{N_L \times N_L}$, we encode it into a more compact conditioning $Y\in \mathbb{M}^{N_Y}$.
Although numerous methods are permissible, we found that pooling can achieve highly salient compression with very little compute cost.
We thus represent the map encoder as a single $N_{pool} \times N_{pool}$ max-pooling layer, with $m_o > m_u > m_f$ and $N_{pool} = N_L / \sqrt{N_Y}$. $Y$ is further one-hot-encoded to capture the discrete nature of $\mathbb{M}$. We use a conditioning size $N_Y=100$, and allow for arbitrary map resolutions $N_L$. In our experiments $N_L=50$.
For efficient array slicing, we extract $L_t$ in world orientation and additionally append the robot angle $\theta_R$ w.r.t. $L_t$ to $Y$.

\textbf{Encoder and Decoder}
The encoder $Q_\phi$ and decoder $P_\psi$ are parametrized as Multi-Layer Perceptrons (MLPs), with 4 layers of 512 hidden units each, using dropout for regularization. They map to a latent representation $z \in \mathbb{R}^3$. 
$Q_\phi$ and $P_\psi$ are then trained jointly to optimize the loss.

\textbf{Loss}
We use $\mathscr{L}$ of Eq.~\ref{eq:elbo} as the loss function. The first term of $\mathscr{L}$ represents the ability of $P_\psi$ to reconstruct the input distribution $X$, which can be modeled as
\begin{equation}
    \mathscr{L}_{rec} = \frac{1}{N_x}\sum_{i=0}^{i=N_x}\lVert P_i - P_i' \rVert^2
    \label{eq:reconstruction_loss}
\end{equation}
where we use the Euclidean norm for $\lVert \{x_i, y_i\} - \{x_i', y_i'\}\rVert$ and the shortest arc for $\lVert \theta_i - \theta_i' \rVert$. 


The second term of $\mathscr{L}$ encourages the input encoding $Z \sim Q_\phi$ to be close to the prior distribution $p(Z|Y) = \mathcal{N}(0,\mathbb{I})$. 
To trade off these objectives, the second term is often weighted by $\lambda_{reg}$, but we empirically found $1$ to work well.

\textbf{Inference}
At test time, it is sufficient to simply draw a latent seed $Z \sim \mathcal{N}(0,\mathbb{I})$ from the prior distribution, and feed it to the decoder with the map conditioning $Y$ to obtain a learned sample.

\subsection{Learning Sample Information Gains}
\label{sec:gain}
The second important component in sampling-based planning is to identify the highest utility sample.
We adopt the formulation of \cite{Schmid20ActivePlanning} and differentiate between total utility $u(P)$, trading off gain $g(P)$ and cost $c(P)$ of a sample $P$.
We employ the common exploration gain formulation: \vspace{-10pt}

\begin{equation}
    g(P) = \sum_{v \in \mathrm{Visible}(P)} f(v), \quad f(v) =
    \begin{cases} 
    1 \text{ if } M(v) = m_u \\
    0 \text{ else}
    \end{cases}
\label{eq:information_gain}
\end{equation}

and use the estimated traversal time $\hat{T}(P)$ as cost $c(P)$. Differentiating gain and cost in this formulation allows our method to be applied to arbitrary cost functions, e.g. for different robot and motion models, without retraining.
The final utility to select the NBV is computed as $u(P) = g(P) / c(P)$ \cite{Schmid20ActivePlanning, Schmid21Glocal}, $\text{NBV} = \arg \max_i \{u(P_i)\}$. Typically, $g(P)$ is computed via ray-casting. We term this method \emph{CVAE}. 
However, since $g(P)$ is oftentimes expensive to compute \cite{Schmid20ActivePlanning}, we propose to directly learn it. For this, we explore three methods.

An intuitive approach is to jointly estimate $P'$ and $g(P')$. We thus directly add $g$ as a coupled target variable to the CVAE data space, i.e. $\bar{X} = \{\bar{P}_i\}_{i=0,\dots,N_X}, \bar{P}_i  = \{x_i, y_i, \theta_i, g_i\}$. We refer to this joint method as \emph{CVAE+GJ}.
However, this approach has the potential limitations of degrading the sampling quality to capture the gain.
Therefore, we also train an independent network $G_\xi(P,Y)$ where $G_\xi$ is an MLP of identical architecture as $Q_\phi$, with parameters $\xi$. To train $G_\xi$, we also add negative, i.e. random, samples to the training set of Sec.~\ref{sec:data} to not bias the estimator. Using this approach, $G_\xi$ can be combined with any sampling method, referred to as \emph{CVAE+G} and \emph{Uniform+G}.

Third, while we found the simple map encoding to work well for the CVAE, for the gain estimation we also present more complex map encoder components, parametrized as a Convolutional Neural Networks (CNN) consisting of three $5 \times 5$ convolutions and a max-pooling layer. These have higher compute cost than the simple encoder, but can result in more complex features. The map encoder is jointly trained with the gain-estimator. We compare this when paired with both the CVAE, and conventional uniform sampling, methods denoted \emph{CVAE+GCNN} and \emph{Uniform+GCNN}, respectively.  


\subsection{Dataset Generation}
\label{sec:data}

\begin{figure}
    \centering
    \includegraphics[width=0.48\columnwidth]{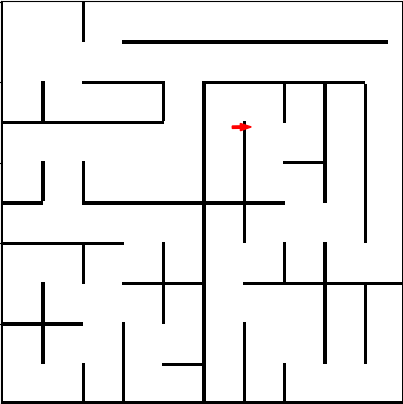}
    \includegraphics[width=0.48\columnwidth]{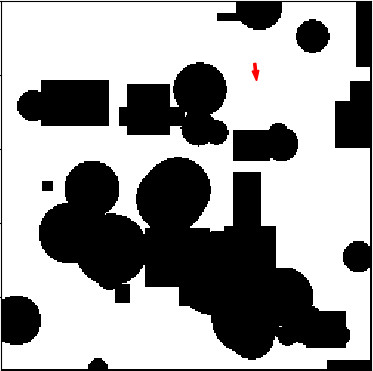}
    \caption{Examples of procedurally generated simulated environments of type \emph{Maze} (left) and \emph{Cluttered} (right). Initial position shown red.}
    \label{fig:simulation_maps}
    \vspace{-20pt}
\end{figure}

To train our networks, a dataset is collected. To achieve a diverse set of environments, simulated worlds are procedurally generated. We implement two types of environments, a classical \emph{Maze} for indoor scenes and \emph{Cluttered} with a lot of fuzzy surfaces, shown in Fig.~\ref{fig:simulation_maps}. The simulator and world generation are made open source with the code$^1$.
Note that only worlds of type \emph{Maze} are used for training.
We use the \emph{Uniform} planner of Sec.~\ref{sec:experimental_setup} to explore each world. For each seen local map, we generate a set of high utility samples $X$ by running the Uniform planner with $N=25$ samples $N_X=20$ times, storing only the NBVs. Examples of this are given in Fig.~\ref{fig:reconstruction} (top row). To optimize $\theta_i$, we further employ orientation optimization \cite{Schmid20ActivePlanning, Selin_nbv_fron, history_nbvp} of each pose.
The total dataset $\mathcal{D} = \{L_i, X_i \}_{i=0,\dots,N_\mathcal{D}}$ contains $N_\mathcal{D}=200k$ samples from $10k$ worlds, $160k$ used for training and $40k$ for validation.

\section{Experimental Setup}
\label{sec:experimental_setup}

\textbf{Simulation}
First, we perform thorough experimental evaluation of all methods in simulation. All methods are run on a separate test set consisting of 5 worlds. For comparability, all methods are run 3 times from identical start poses. The mean and standard deviation over these 15 runs is reported. An omni-directional robot with linear and angular velocities of $v_{max}=1 m/s$ and $\Omega_{max}=1 rad/s$, equipped with an RGB-D camera with $FoV=90 ^\circ$ and range $r=5 m$, is modelled. We use a local map of twice the sensing range as suggested in \cite{Schmid21Glocal}.

\textbf{Baselines}
We compare our proposed methods against the following set of baselines. 
\emph{Uniform} is a classical sampling-based planner, that uses uniform random sampling to generate $N$ reachable poses $P_i$, computes $g(P_i)$ via ray-casting, and chooses $P_t = \arg \max_{i} u(P_i)$. T
We further compare against the established method of \cite{nbvp_bircher}, termed \emph{RH-NBVP}.
As a learning-based method, \emph{Imitation} directly predicts the one best $P_t$ from the map conditioning on $Y$. This is a conventional behavior cloning approach using regression on the targets with square loss \cite{Bai2017}. To this end, we trained a separate network $R_\varphi(Y)$ with identical architecture to $Q_\phi$ on our dataset.
We also tried to train an RL baseline based on GA3C \cite{babaeizadeh2017ga3c} like in \cite{zhu2018}. However, RL is known to be difficult to train \cite{henderson2018deep} and after considerable effort to tune parameters, architecture, and rewards, no presentable results were achieved. While RL holds great promise, it appears our CVAE-based approach is easier to train, at least on our problem with more complex robot sensor and motion models than in related work on RL.

\textbf{Hardware}
Simulation experiments are conducted on an AMD Epyc 7742 CPU @2.25GHz. All methods are run single-threaded to allow a fair comparison.

\textbf{Robot Experiments}
We further validate our method on a TurtleBot3 
(TB) 'Burger', with $v_{max}=8 cm/s$ and $\Omega_{max} = 0.3 rad/s$. It is equipped with an Intel Realsense D435
RGB-D camera of $FoV=87^\circ$ and $r=2.5m$. 
We use the standard TB stack for online state estimation, a custom PD controller to track trajectories in narrow environments, and the Single-TSDF implementation of \cite{schmid2021panoptic} to extract safe local maps $L_t$ from the depth camera.
All computation is performed on a laptop with an Intel i7-8550U CPU @1.80GHz.

\section{Results}
\label{sec:results}


\subsection{Distribution and Gain Learning}

\begin{figure}[tb]
    \centering
     \includegraphics[width=0.32\columnwidth]{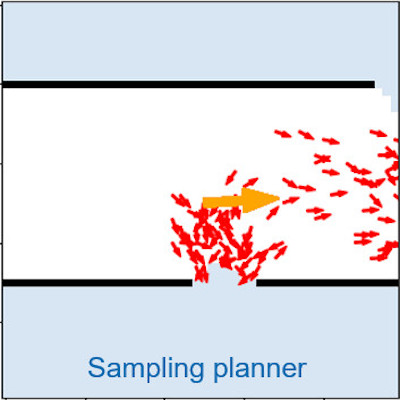}
     \includegraphics[width=0.32\columnwidth]{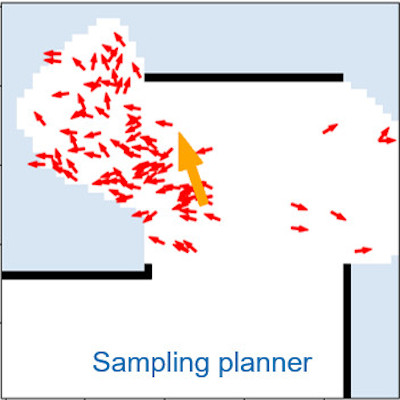} 
     \includegraphics[width=0.32\columnwidth]{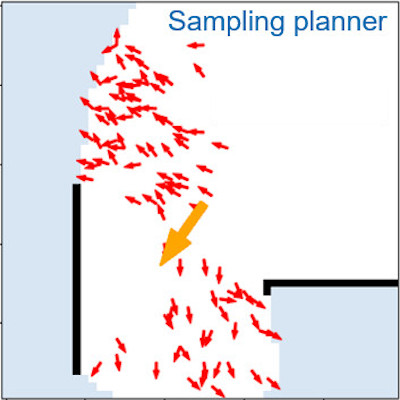}  \\ \vspace{3pt}
    \includegraphics[width=0.32\columnwidth]{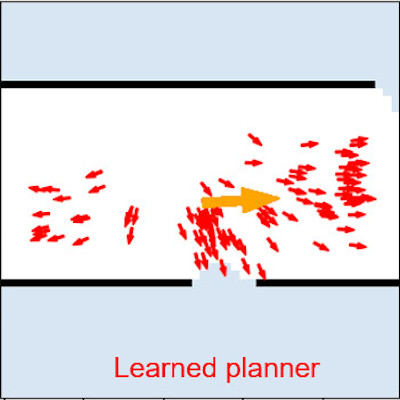} 
    \includegraphics[width=0.32\columnwidth]{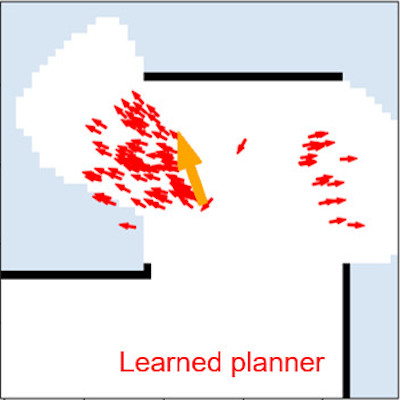} 
    \includegraphics[width=0.32\columnwidth]{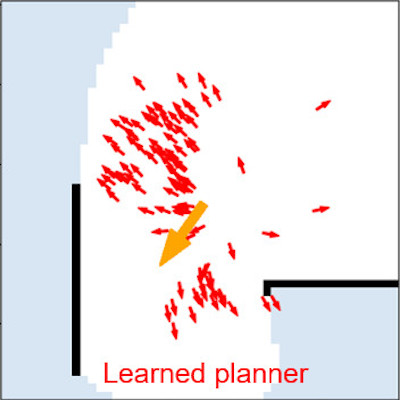}
    \caption{Qualitative examples of the learnt distributions. Top: Training maps, each arrow represents the best action of an independent run of the teacher planner. Bottom: Individual samples drawn from our learnt distribution.}
    \label{fig:reconstruction}
    \vspace{-10pt}
\end{figure}

\begin{table}[tb]
    \centering
    \caption{Average true utility of chosen actions by sampling method, gain computation method, and number of samples $N$.}
    \vspace{-5pt}
    \begin{adjustbox}{max width=\columnwidth}
    \begin{tabular}{l|c|ccc|ccc}
    & \multirow{2}{*}{$N=1$} & \multicolumn{3}{c}{$N=10$} & \multicolumn{3}{|c}{$N=50$} \\
    Samples from & & Ray & +G & +GCNN & Ray & +G & +GCNN \\
    \midrule
    Training Data & 63.1 & 80.1 & 69.2 & 70.1 & 87.3 & 68.7 & 70.8 \\
    CVAE (ours) & 51.2 & 75.3 & 49.1 & 62.5 & 80.7 & 42.3 & 62.4 \\
    Uniform & 10.5 & 43.3 & 20.5 & 35.0 & 62.9 & 22.2 & 47.3 \\
    \end{tabular}
    \end{adjustbox}
    \label{tab:gain}
    \vspace{-20pt}
\end{table}

First, we qualitatively analyze the performance of the proposed distribution learning method, shown in Fig.~\ref{fig:reconstruction}.
The wide spread of NBVs highlights the high variance and multi-modality in sampling-based exploration planning.
We observe that our method is well able to capture this multi-modality and produces highly informative samples. This is also reflected in Tab.~\ref{tab:gain} showing the utility of chosen actions. Our method produces actions close to the training data, showing a fivefold improvement over Uniform for low $N$. As $N$ and therefore coverage increases, the choice of gain estimator becomes more important.


\subsection{Exploration Performance}
\label{sec:exploration_performance}

\begin{figure*}[tb]
    \centering
    \includegraphics[width=\linewidth]{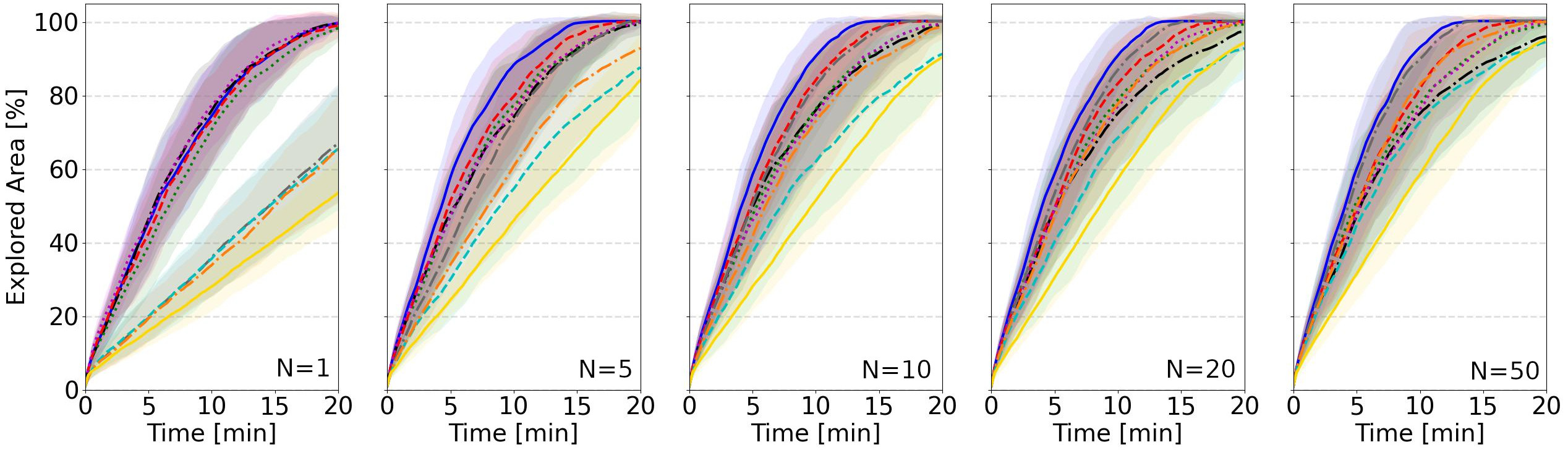} \vspace{-13pt} \\
    \includegraphics[width=\linewidth]{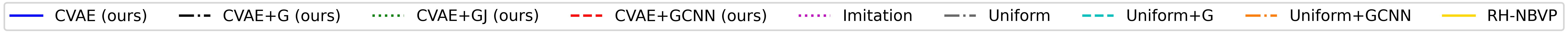}
    \caption{Exploration performance in the \emph{Maze} environment for numbers of samples $N=\{1,5,10,20,50\}$. Methods are labeled as in Sec.~\ref{sec:gain} by sampling scheme (\emph{CVAE}, \emph{Uniform}) and gain computation type, for ray-casting (\emph{blank}), jointly estimated (\emph{+GJ}), pooling-based (+G) and CNN-based (\emph{+GCNN}), respectively. We observe that the learnt sampling distribution significantly improves performance, most notably for few samples. As $N$ and therewith coverage increases, the information gain computation becomes more important.  }
    \label{fig:exploration}
    \vspace{-15pt}
\end{figure*}

To evaluate exploration performance, the exploration progress for each method, using $N=\{1,5,10,20,50\}$ samples per planning step, is shown in Fig.~\ref{fig:exploration}. The time and distance until exploration is complete is shown in Tab.~\ref{tab:exploration}.

For $N=1$ the importance of the learnt sampling distribution is evident. By drawing a single sample from the model, the CVAE-based methods show significantly higher performance than their classical counterparts.
As the number of samples increases, CVAE further improves in performance, reducing the time until exploration is completed by $42$\% at 50 samples.
Naturally, as the coverage becomes more complete for increasing $N$, the influence of the sampling distribution decreases.
Notably, even for 50 samples, after which no significant performance improvements were found, CVAE still shows similar or better performance than uniform.
This suggests that the optimal viewpoint is contained within the learned distribution.

The importance of the gain computation becomes evident as the number of samples increases. Here, the ray-casting based methods act as the upper bound since they compute the true gain, reflected in the high performance of CVAE and Uniform. 
Importantly, methods that learn an information gain improve with more samples being drawn, with the exception of CVAE+G. 
This is explained by the high variance of the $\max(\cdot)$ operator, where a single noisy or overfitted prediction can decrease exploration performance, as is also observed in Tab.~\ref{tab:gain}. 
In contrast to this, the CNN-based gain estimator is able to better predict the information gains, leading to improved performance for more samples with CVAE+GCNN. The trend is even more pronounced in Uniform+GCNN. 
This behavior for CVAE+G and CVAE+GCNN highlights the interplay between sampling and gain evaluation and opens interesting avenues for future research. The joint architecture in CVAE+GJ has to trade off learning informed sampling and accurate gains at the same time and is not able to improve performance.
RH-NBVP exhibits slightly lower performance than the other methods. This can be explained by the sub-optimal wiring of its RRT, whereas the other methods including Uniform follow the approach of \cite{Schmid21Glocal} and plan for a short horizon to incorporate the new information.

We observe that the distribution learning methods improve with more samples being added, compared to imitation learning. This suggests that the multi-modal nature of sampling translates to increased exploration performance. In addition, Imitation is the only method that failed to produce a feasible path for 2.5\% cases.

The exploration time is primarily governed by the quality of the selected view points, which also manifests in the travel distances shown in Tab.~\ref{tab:exploration}. The only exception is RH-NBVP, which does not account for the robot's dynamics and chooses shorter but slower paths.

\begin{table}[tb]
    \centering
    \caption{Mean time to complete exploration [min] (left) and total travel distance [m] (right). A dash indicates incomplete exploration.}
    \begin{adjustbox}{max width=\columnwidth}
    \vspace{-5pt}
    \begin{tabular}{lrlrlrl}
     Method & \multicolumn{2}{c}{$N=1$} & \multicolumn{2}{c}{$N=10$} & \multicolumn{2}{c}{$N=50$} \\
    \midrule
  CVAE (ours) & \textbf{22.8} & 4728 &\textbf{15.0} & \textbf{3071} & \textbf{13.2} & \textbf{2663} \\
CVAE+G (ours) & \textbf{22.8} & 4703 & 24.9 & 5361 & 32.3 & 6724  \\
CVAE+GJ (ours) & 24.6 & 5246 & 22.8 & 5379 & 23.1 & 5500\\
CVAE+GCNN  & 22.9 & \textbf{4720} & 19.4 & 4162 & 19.9 & 4145 \\
Imitation  & 25.2 & 4981 & 25.2 & 4981& 25.2 & 4981 \\
Uniform    & \multicolumn{2}{c}{-}  & 16.2 & 3616 & 13.5 & 2920 \\
Uniform+G  & \multicolumn{2}{c}{-} & 28.9 & 6895 & 30.0 & 7144 \\
Uniform+CNN& \multicolumn{2}{c}{-} & 21.9 & 5049 & 19.3 & 4405 \\
RH-NBVP    & \multicolumn{2}{c}{-} & 28.8 & 3381 & 24.9 & 2935 \\
    \end{tabular}
     \end{adjustbox}
    \label{tab:exploration}
    \vspace{-10pt}
\end{table}


\subsection{Generalization}

\begin{figure}[tb]
    \centering
    \includegraphics[width=0.9\linewidth]{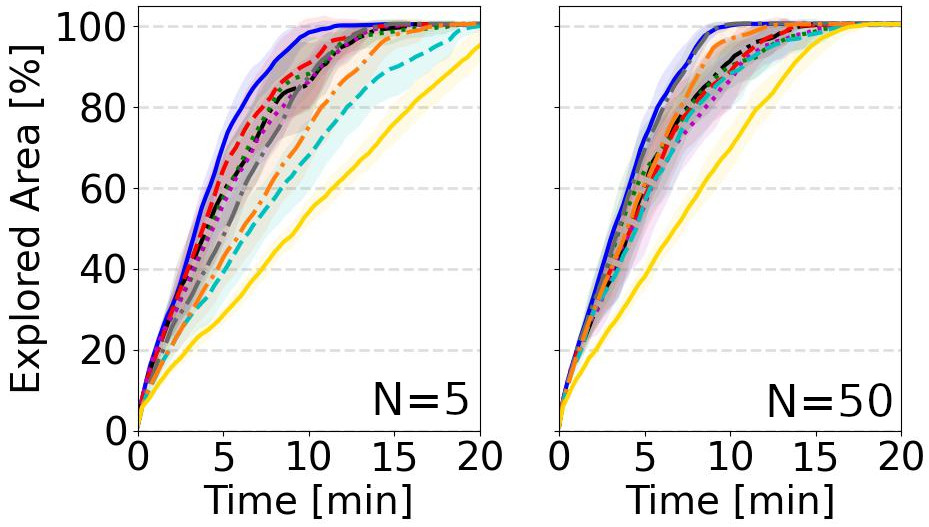}
    \caption{Generalization performance evaluated in \emph{Cluttered} environments. Note that all models were only trained on \emph{Maze} data. We find that the performance is highly similar to the Maze environment, suggesting our method generalizes well to other environments.}
    \label{fig:generalization}
    \vspace{-15pt}
\end{figure}

To evaluate the generalization performance of our method, we conduct experiments in \emph{Cluttered} environments. Note that all models were only trained on \emph{Maze} data, which is distinctly different as shown in Fig.~\ref{fig:simulation_maps}. 
The results for $N=\{5,50\}$ are presented in Fig.~\ref{fig:generalization}.

Even in this novel environment, we find that the performance is highly similar to the Maze environment. Again, CVAE outperforms ($N=5$) or matches ($N=50$) the other methods. Similarly, all the learned gain models are able to identify informative views and show strong performance.
This corroborates the findings of Sec.~\ref{sec:exploration_performance} and suggests that both the learnt sampling-distribution and gain prediction generalize to different environments.


\subsection{Computation Cost}

\begin{figure}[tb]
    \centering
    \includegraphics[width=\linewidth]{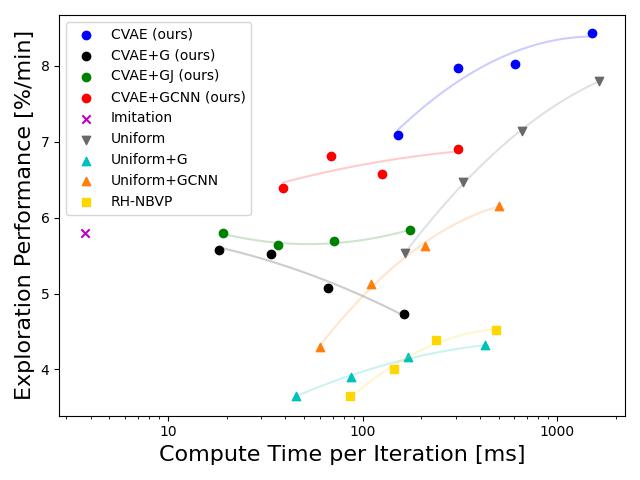}
    \vspace{-20pt}
    \caption{Pareto plot of performance vs. compute cost. Performance is measured by the time needed to explore 90\% of the maze, which is the highest score all planners reached. Each $N \in \{5,10,20,50\}$ is represented by a data point. Lines indicate a fitted trend. We observe that the CVAE is the dominant option for the high compute regime, whereas CVAE+GCNN is a favorable alternative for computationally constrained platforms.}
    \label{fig:compute}
    \vspace{-20pt}
\end{figure}

Computing power is a crucial and limited resource on a mobile robot. Depending on hardware limitations, energy constraints, and the other tasks the robot has to perform, different weights of the computation objective $\gamma$ (Sec.~\ref{sec:problem}) are desirable. 
This compute vs. performance trade-off for all methods is presented as a Pareto curve in Fig.~\ref{fig:compute}.

Clearly, the raycasting-based gain methods show the highest performance. However, the explicit gain calculation comes at significant computational cost. 
Here, CVAE offers a strictly superior option to the classical planner, improving performance by up to $28$\% at similar cost, or achieving similar performance while reducing computation $5.3$ times.

The advantage of learning the information gain is apparent in the significantly reduced compute cost. CVAE+GCNN exhibits the best performance in the low-compute area, although the performance gain for more samples is less pronounced. 
Similar behavior is also apparent for random sampling.  While Uniform+G/+GCNN benefit from more samples, they only offer improved performance in the low-cost regime. As expected, Imitation offers a very fast but rather poor result.

\subsection{Robot Experiments}
\label{sec:robot_experiments}

\begin{figure}[tb]
    \centering
    \includegraphics[width=0.32\columnwidth]{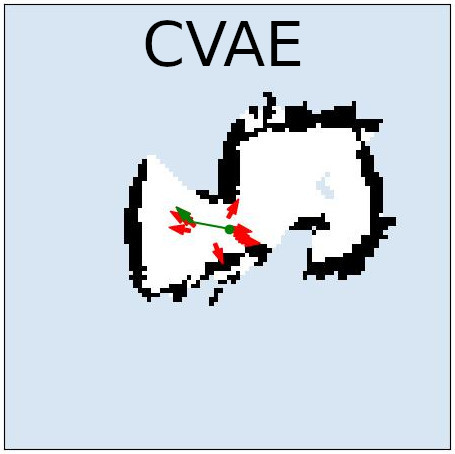}
    \includegraphics[width=0.32\columnwidth]{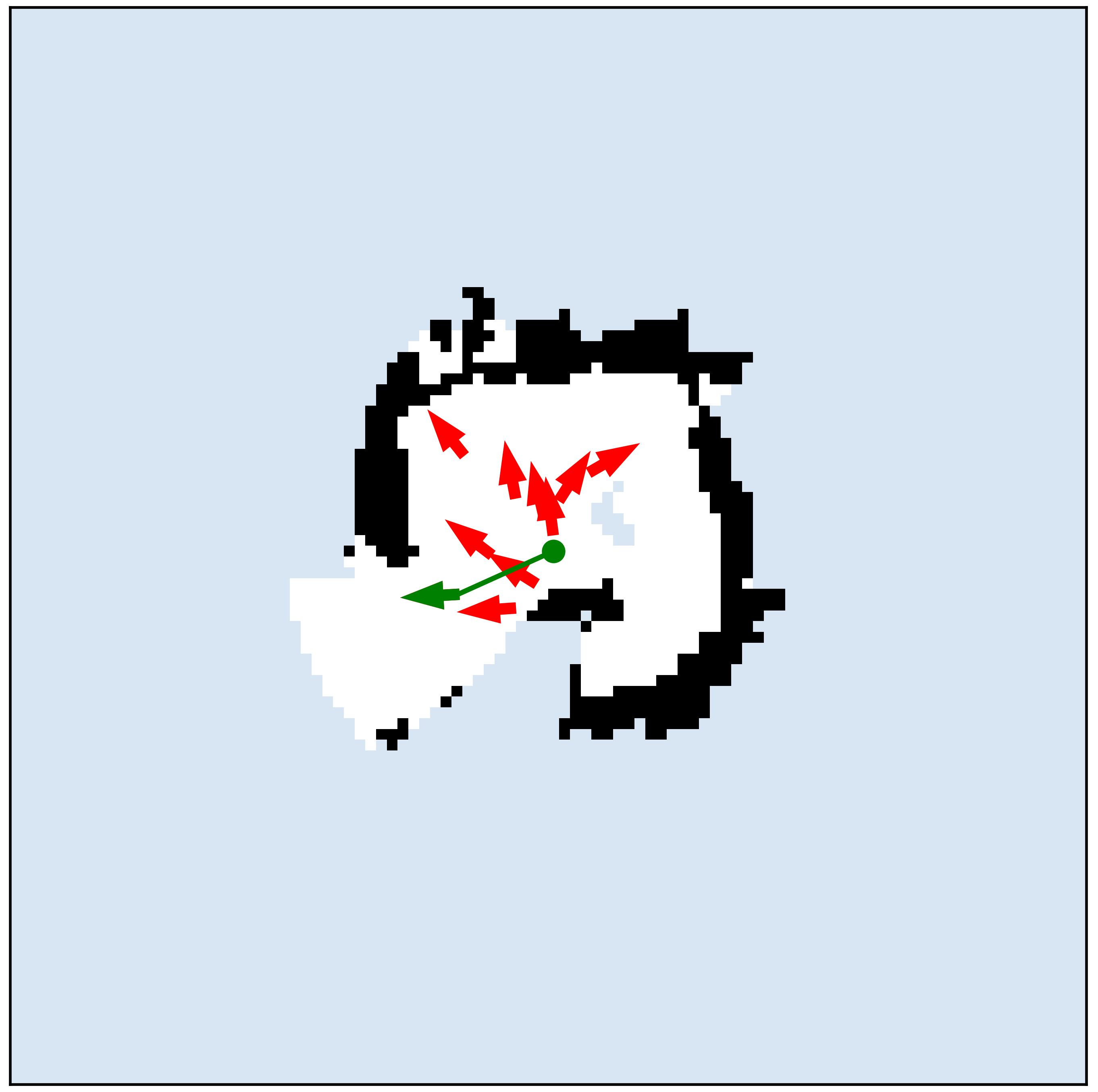}  
    \includegraphics[width=0.32\columnwidth]{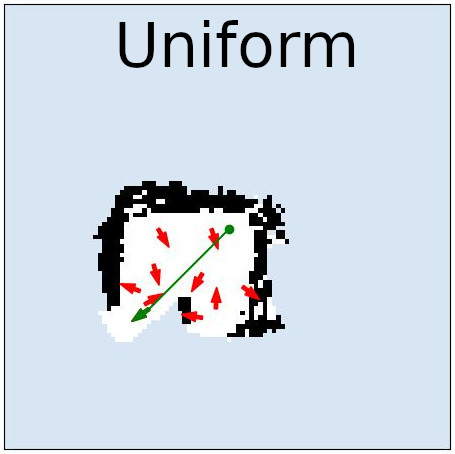}  
    \\ \vspace{3pt}     \includegraphics[width=0.32\columnwidth]{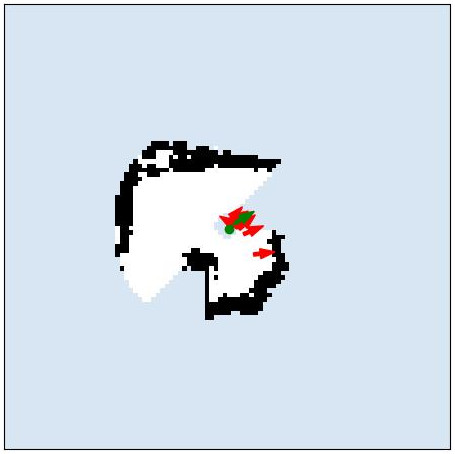}
    \includegraphics[width=0.32\columnwidth]{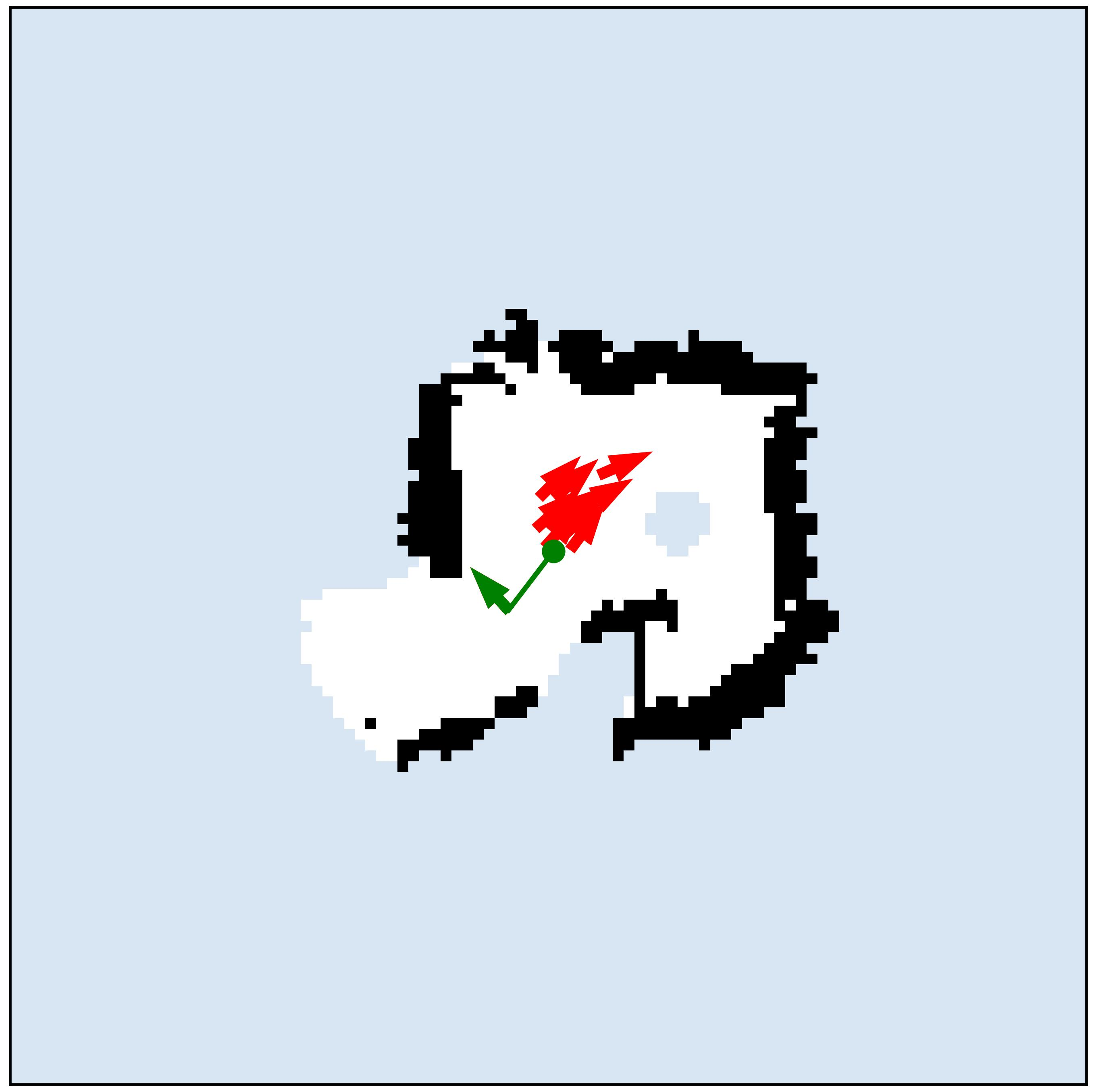} 
    \includegraphics[width=0.32\columnwidth]{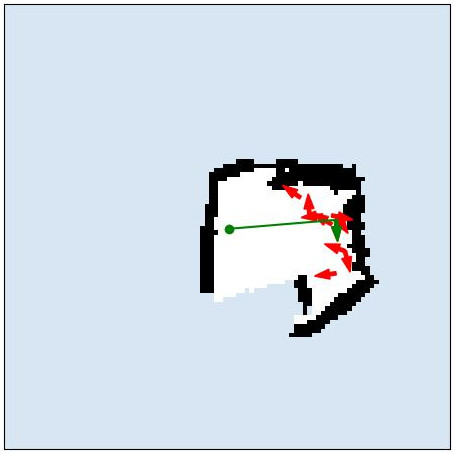}
    \caption{Selected local maps $L_t$, with drawn samples (red arrows) and selected action (green) during the robot experiments. Even on this noisy real map, our CVAE is able to capture the bi-modal options of going either left or right (top-left) or identify the best single best move (bottom-left).
    Potential limitations of the approach are highlighted in the center column.
   In some cases, the method is not able to predict an adequate distribution. However, it falls back to good uniform coverage in that case(top). Rarely, the method fails and predicts an erroneous distribution (bottom). The uniform planner exhibits high variance, as sometimes close to perfect moves are generated (top-right), at other times none of the samples are useful (bottom-right).}
    \label{fig:tb_qualitative}
    \vspace{-10pt}
\end{figure}

\begin{figure}[tb]
    \centering
    \includegraphics[width=\linewidth]{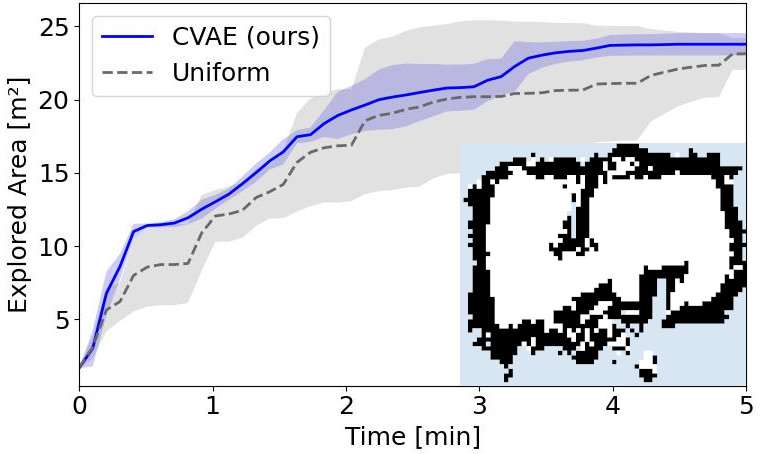} 
    \vspace{-15pt}
    \caption{Quantitative evaluation on the mobile robot over 3 runs. The bottom right shows a final map after CVAE exploration.}
    \label{fig:tb_exploration}
    \vspace{-20pt}
\end{figure}

\begin{figure}[tb]
    \centering
    \includegraphics[height=0.5\linewidth]{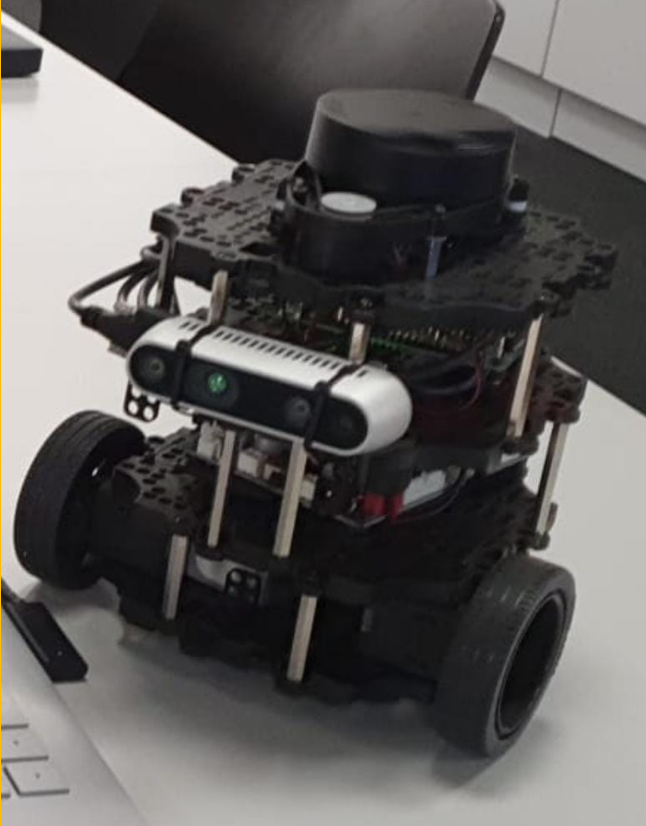} 
    \includegraphics[height=0.5\linewidth]{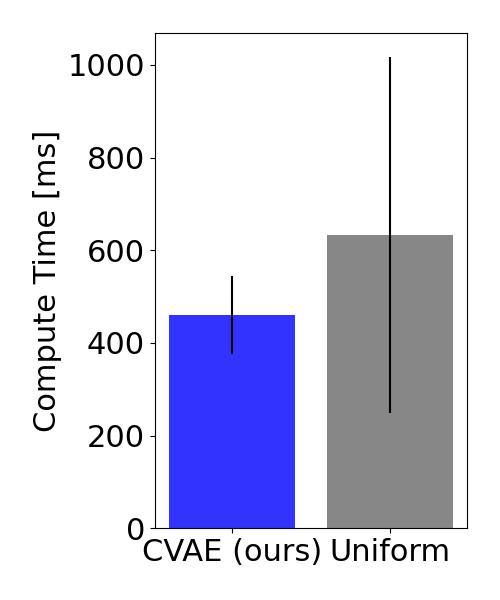} 
    \caption{Overview of the employed robot (left). Compute time consumed per iteration during the robot experiment (right).}
    \label{fig:tb_robot}
    \vspace{-20pt}
\end{figure}

To further validate our approach, experiments on a low-cost mobile robot are conducted. We employ a Turtlebot equipped with a realsense D435 depth camera, shown in Fig.~\ref{fig:tb_robot}, to explore a challenging, cluttered indoor environment, shown in Fig.~\ref{fig:intro_showcase}.
Three experiments of CVAE and Uniform for $N=10$ are performed.

Fig.~\ref{fig:tb_qualitative} shows example local maps, drawn samples, and selected NBVs from the real system. 
Due to the high sensor noise and localization errors, the map extracted from \cite{schmid2021panoptic} after depth fusion is notably more noisy than in simulation. Together with the different scale of the narrow environment, this constitutes a major domain shift.
Nonetheless, our network purely trained in simulation is able to generate meaningful samples most of the time.
We observe that the Uniform planner exhibits high-variance, sometimes generating close to optimal views, at other times missing the obvious.

This is also reflected in the quantitative results reported in Fig.~\ref{fig:tb_exploration}. While both methods are able to successfully explore the whole environment, CVAE tends to show quicker and more consistent exploration.
A final map obtained after exploration is shown in Fig.~\ref{fig:tb_exploration} bottom. While the map quality primarily depends on the localization and sensing noise, we observe that our planner thoroughly explores all free space.
Lastly, Fig.~\ref{fig:tb_robot}, right, shows the computation time per planning step. CVAE shows improved computation cost, as many more random samples need to be checked to achieve 10 feasible poses in this complex environment.

\section{Conclusions}
\label{sec:conclusion}

In this work, we consider the problem of learning sampling-based local exploration planning.
We propose a CVAE-based approach to directly predict an, often multi-modal, informed sample distribution directly from standard partially-observed occupancy maps. 
We show in thorough experimental evaluation that the presented approach can either improve exploration performance by up to $28$\% or match performance and reduce computation by a factor of five compared to conventional approaches. Furthermore, we explore a variety of methods to combine sampling and gain computation. We find that learning the information gain with a CNN map encoder, in combination with the CVAE-based sampling strategy, leads to favorable performance vs. compute trade-offs for compute constrained systems. We show both in simulation and on a low-cost mobile robot that our method generalizes well to different environments.




{\small
\bibliographystyle{IEEEtran}
\bibliography{IEEEfull,references}
}


\end{document}